\documentclass[10pt,twocolumn,letterpaper]{article}

\usepackage{cvpr}              
\usepackage{multirow}

\definecolor{cvprblue}{rgb}{0.21,0.49,0.74}
\usepackage[pagebackref,breaklinks,colorlinks,allcolors=cvprblue]{hyperref}

\def\paperID{6} 
\def\confName{CVPR}
\def\confYear{2026}

\title{SLU-2K: A Question-Based Benchmark for Semantic Evaluation of Sign Language Translation}

\author{Zeno Testa\\
University of Modena and Reggio Emilia\\
Modena, Italy\\
{\tt\small 357209@studenti.unimore.it}
\and
Antonino Furnari\\
University of Catania\\
Catania, Italy\\
{\tt\small antonino.furnari@unict.it}
\and
Lorenzo Baraldi\\
University of Modena and Reggio Emilia\\
Modena, Italy\\
{\tt\small lorenzo.baraldi@unimore.it}
\and
Natalia Díaz-Rodríguez\\
University of Granada, \\
CITIC \& DaSCI Institute, Spain\\
{\tt\small nataliadiaz@ugr.es}
}

\begin{document}

\maketitle


\begin{abstract}
Sign Language Translation (SLT) is typically evaluated with surface-form metrics such as BLEU and ROUGE, which reward lexical overlap but do not directly measure whether a translation preserves the meaning of the source sign sequence. 
This is in contrast with the final objective of integrating SLT in assistive technology. 
In this work, we shift the focus from Sign Language Translation (SLT) to Sign Language Understanding (SLU), with particular emphasis on semantic understanding. Specifically, we evaluate systems based on their ability to correctly recover, from the input video, key semantic aspects of the original sentence, such as actions taking place and facts about people and objects.
To enable this evaluation systematically, we propose SLU-2K, a dataset of 2,350 closed-ended video question-answer pairs based on the popular PHOENIX-2014T and CSL-Daily datasets.
To obtain SLU-2K, we propose and extensively evaluate an automated data generation pipeline which produces questions across $7$ categories, namely \textit{actions}, \textit{locations}, \textit{numbers}, \textit{objects}, \textit{people}, \textit{time}, and \textit{weather conditions}.
We show the potential of SLU-2K by evaluating popular Multimodal Large Language Models (MLLMs) and two representative state-of-the-art systems, MMSTL and SpaMo. 
Our results show that MLLMs reach near-random performance, highlighting the need for a more systematic integration of SLU in current AI systems. Furthermore, state-of-the-art translation systems carefully fine-tuned on in-domain data still exhibit a substantial semantic gap, with results ranging from $56.7\%$ to $75.2\%$.
These findings suggest that current SLT evaluation protocols overestimate true understanding and that future progress should be measured not only by fluency and n-gram overlap, but also by semantic correctness. Code, prompts, and benchmark files are available at \url{https://github.com/ZenoTsT/SLU-2K}
\end{abstract}


\section{Introduction}
\label{sec:intro}
Sign Language Translation (SLT) aims to translate visual sign language input into spoken-language text. In recent years, the field has progressed substantially thanks to stronger visual encoders, multimodal fusion architectures, and large sequence-to-sequence models~\cite{voskou2021stochastic,zhang2023sltunet,zhou2023gloss,gong2024llms,hwang2024efficient,kim2025leveraging,tan2025multilingual,jang2025lost,asasi2025hierarchical,chen2024factorized,chen2025c,asasi2025beyond,hwang2025spatio,wong2024sign2gpt,yeo2025towards}. Despite these improvements, evaluation in SLT is still dominated by text-overlap metrics such as BLEU and ROUGE. These metrics remain useful for measuring lexical similarity, but they are poor proxies for semantic faithfulness. A translation can obtain a reasonable overlap score while still changing a key number, replacing one location with another, or altering the timing of an event.
\begin{figure}
    \centering
    \includegraphics[width=\linewidth]
    {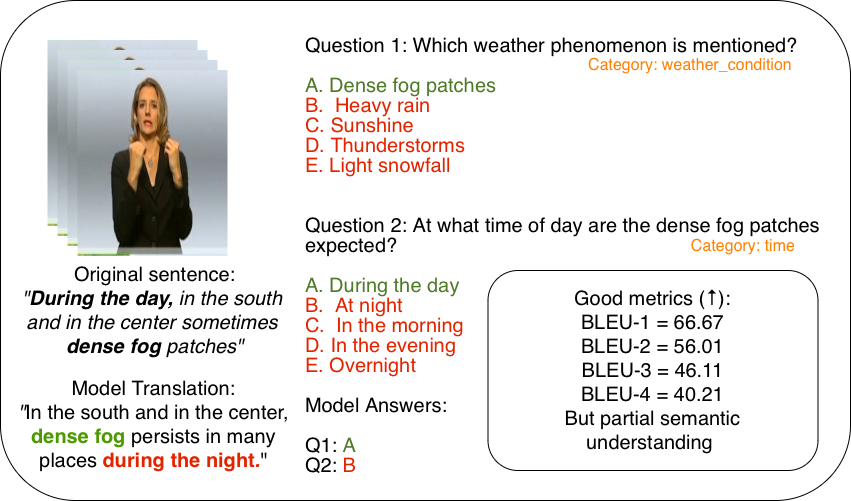}
    \caption{The proposed Sign Language Understanding SLU-2K benchmark aims to assess understanding beyond literal translation. Despite the good BLEU scores due to lexical overlap, the model wrongly predicts “B. At night” for question two, due to an imprecise translation.}
    \label{fig:teaser}
\end{figure}
As the final goal of these efforts would be to integrate SLT systems in assistive technology (\textit{e.g.}, smart glasses), we argue that semantic understanding should be evaluated directly, rather than relying on mere lexical overlap.
Indeed, semantic preservation is exactly what many real-world applications require. For example, in weather broadcasts, a mistranslated temperature, date, or location can materially alter meaning. 

Motivated by this need, we frame the problem from a Sign Language Understanding (SLU) perspective, where systems are evaluated by their ability to understand the content of the sign video, rather than just providing a translation.
Instead of scoring a model only by n-gram overlap against a reference translation, we derive targeted multiple-choice questions from the reference translation and evaluate whether the model’s predicted translation contains sufficient information to answer them correctly. To help evaluation, we propose SLU-2K a benchmark of $2,350$ question-answer sets automatically generated from the PHOENIX~\cite{camgoz2018neural} and CSL-Daily~\cite{zhou2021improving} datasets and
grouped into interpretable semantic categories such as \emph{location}, \emph{numbers}, \emph{time}, and \emph{weather condition} for PHOENIX, and \emph{action}, \emph{person}, \emph{object}, \emph{location}, \emph{time}, and \emph{numbers} for CSL-Daily (see Fig.~\ref{fig:teaser}). This formulation enables a fine-grained analysis of which semantic elements are preserved and which are systematically degraded.

A central contribution of our work is not only the final benchmark itself, but also the automatic pipeline used to construct it. We generate questions and distractors with a reliable LLM, then subject them to a multi-stage filtering process that removes weak or ambiguous items. At the end, the surviving questions are checked for answerability when the original sentence is provided. This process is designed to ensure that retained questions are both non-ambiguous and semantically grounded.

Our benchmark is particularly motivated by two practical observations. First, standard SLT benchmarks differ greatly in linguistic complexity. PHOENIX-2014T contains weather-forecast style utterances with restricted semantics and recurring syntactic patterns, while CSL-Daily covers broader and less repetitive content. Second, recent SLT models often optimize directly or indirectly for textual generation quality, which may favour fluency more than factual precision. A model may produce well-formed text that is globally plausible yet locally incorrect in semantically important details.

We evaluate our benchmark on one open source and one proprietary Multimodal Large Language model, as well as on translations from two representative systems, MMSTL~\cite{kim2025leveraging} and SpaMo~\cite{hwang2024efficient}. The results reveal clear and actionable patterns. First, Multimodal Large Language models achieve near-random performance, highlighting the need for a more systematic integration and evaluation of SLU abilities in these models, for which we believe our benchmark will be a useful tool.
Second, even state-of-the-art translation systems fine-tuned on the target datasets achieve limited performance. On PHOENIX, MMSTL achieves 75.2\% overall semantic accuracy, while SpaMo reaches 68.5\%. The largest gap appears in the \emph{numbers} category, suggesting that exact-value preservation remains a major challenge. On CSL-Daily, MMSTL reaches 56.7\%, substantially lower than on PHOENIX, which indicates that semantic evaluation becomes harder as the language becomes less syntactically regular and semantically richer.

These results support the central claim of this paper: \emph{surface-level translation quality is not the same as semantic understanding}. A system can appear strong under conventional metrics while still failing to preserve fine-grained meaning. Our framework is designed to expose this gap and to provide a complementary evaluation protocol that is interpretable, category-aware, and directly tied to semantic correctness.

\noindent Our main contributions are threefold:

    1) We introduce a question-based benchmark framework for semantic evaluation of SLT, and instantiate it on PHOENIX-2014T and CSL-Daily, providing category-level semantic benchmarks for two widely used datasets. 
    
    
    2) We propose a reusable automatic pipeline for generating, filtering and validating semantic multiple-choice questions from SLT reference sentences.
    
    3) We show experimentally that strong SLT systems still suffer from substantial semantic errors, especially on precise factual content such as numbers, locations, and temporal expressions.

\section{Related Work}
\label{sec:related}

\subsection{Sign Language Translation}

SLT has evolved from early gloss-based pipelines to end-to-end visual-to-text architectures. Earlier methods often relied on gloss supervision as an intermediate representation ~\cite{camgoz2018neural,zhang2023sltunet,mercanoglu2025spotter+}, while more recent systems increasingly seek gloss-free or weakly supervised alternatives. Transformer-based sequence models, pretrained language models, and large-scale visual encoders have all contributed to improved translation quality~\cite{kim2025leveraging, gong2024llms, wong2024sign2gpt, chen2024factorized, asasi2025beyond, jang2025lost, yeo2025towards}. Recent systems have also explored multimodal fusion between spatial and temporal features, as well as stronger decoding strategies for natural-language generation \cite{hwang2024efficient, hwang2025spatio}.

Although the modeling landscape has advanced considerably, the dominant evaluation protocol has remained largely unchanged. Most works still report BLEU~\cite{papineni2002bleu} and ROUGE~\cite{lin2004rouge} as primary indicators of translation quality. These metrics are useful for comparison with earlier literature, but they are limited in their ability to capture semantic equivalence. This limitation is especially important in SLT, where generated text may be fluent while failing to preserve one or more crucial source facts.

\subsection{Evaluation Beyond Surface Overlap}

The mismatch between lexical overlap and semantic correctness is not unique to SLT. In machine translation, summarization, and image captioning, researchers have long noted that n-gram metrics reward surface similarity more than factual fidelity ~\cite{li2022faithfulness}. More recent work in natural language generation has explored question answering, entailment, and factual consistency as alternatives or complements to overlap-based evaluation ~\cite{honovich2021q2,honovich2022true}.

Our work brings a similar perspective to SLT. Rather than asking whether a prediction is close to a reference at the token level, we ask whether the prediction preserves the answer to targeted semantic questions derived from the reference. This design yields two advantages. First, it makes evaluation interpretable: one can directly inspect which semantic categories fail. Second, it better aligns with the intended function of SLT, namely meaning transfer.

\begin{figure*}[t]
    \centering
    \includegraphics[width=0.95\textwidth]{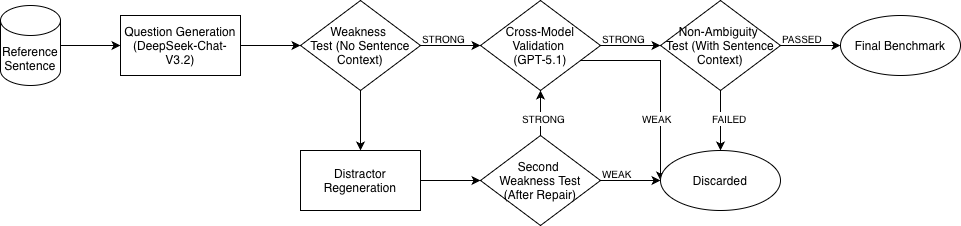}
    \caption{Overview of the SLU-2K Benchmark construction pipeline. A question and its distractors are first generated from a reference sentence to be translated. The item is then subjected to multiple validation stages: weak questions undergo distractor regeneration, strong candidates are cross-validated with a second LLM, and all surviving items are finally checked for non-ambiguity by verifying that the original sentence is sufficient to answer the question correctly. Only questions that pass all stages are added to the SLU-2K benchmark.}
    \label{fig:pipeline}
\end{figure*}

\subsection{Dataset Complexity and Semantic Evaluation}

Benchmark difficulty in SLT depends not only on visual quality and sequence length, but also on linguistic and semantic diversity. PHOENIX-2014T ~\cite{camgoz2018neural} is valuable and widely adopted, yet its domain is relatively narrow and repetitive. CSL-Daily ~\cite{zhou2021improving}, in contrast, covers broader communicative content and can therefore stress semantic evaluation more strongly. This distinction is central to our analysis: if semantic accuracy drops sharply across datasets, then evaluation protocols must account for dataset-specific complexity rather than assuming that performance transfers uniformly.

\section{SLU-2K Benchmark Construction}
\label{sec:benchmark}

We design SLU-2K benchmark to measure whether a model prediction preserves semantically relevant information from a reference sentence. Rather than comparing the generated and ground truth translations holistically, we decompose each sentence into a set of answerable semantic units. Each unit is instantiated as a multiple-choice question. A model translation is considered correct for a given item if it supports the same answer as the reference.
This formulation turns semantic evaluation into a structured comprehension task. If a translation changes a city name, a date, a number, or the described weather condition, the corresponding question can expose the error even when the rest of the sentence remains fluent.

\subsection{Overview of the Data Generation Pipeline}

Our benchmark construction process is fully automatic and is designed to retain only questions that are semantically meaningful, sufficiently discriminative, and non-ambiguous. The full pipeline is illustrated in \cref{fig:pipeline}. Starting from a reference sentence (a ground truth translation from a SLT dataset), we first generate a category-specific multiple-choice question together with one correct answer and four distractors. The generated item is then validated through several consecutive tests, including distractor repair and cross-model checking, before being included into the final benchmark.

A key design principle is that a benchmark question should \emph{require access to the sentence semantics}. If a language model can answer a question correctly without seeing the original sentence, then the item is weak: it can be solved using commonsense priors, frequency biases, or superficial plausibility rather than from the underlying semantic content. Our pipeline explicitly detects and filters such cases.

\subsection{Question Generation}

For each reference sentence that is to be translated, we generate category-aware multiple-choice questions using a reliable instruction-following LLM, namely DeepSeek-Chat-V3.2. The model receives a prompt tailored to the target semantic category and is asked to produce one question grounded in the sentence, one correct answer, and four distractors.

The targeted categories are dataset dependent. For PHOENIX, which is focused on weather forecasts, we generate questions about \emph{location}, \emph{time}, \emph{numbers}, and \emph{weather condition}. For CSL-Daily, we target \emph{action}, \emph{person}, \emph{object}, \emph{location}, \emph{time}, and \emph{numbers}. This category-aware design allows the benchmark to focus on semantically meaningful units that are both interpretable and diagnostically useful.

\subsection{Weakness Test Without Sentence Context}

After generation, each question is tested to determine whether it can be answered \emph{without access to the original sentence}. In this stage, the LLM (DeepSeek-Chat-V3.2) receives only the question and the candidate answers, but not the sentence from which the item was constructed.

If the model is able to answer correctly in this setting, the question is classified as \textbf{weak}. Such an item is not sufficiently diagnostic, because it can be solved without relying on sentence-specific semantics. If, instead, the model fails to answer correctly, the question is classified as \textbf{strong}, since it appears to require actual semantic information from the source sentence.

\subsection{Distractor Regeneration and 2\textsuperscript{nd} Weakness Test}

Questions classified as weak in the first test are not discarded immediately. Instead, their distractors are regenerated in an attempt to make the item harder and more discriminative. This \emph{distractor strengthening} stage keeps the original semantic focus of the question while replacing the answer options with alternatives that are more competitive.

Once the distractors have been regenerated, the question is tested again under the same condition as before: the LLM (DeepSeek-Chat-V3.2) receives the question and answer choices, but not the original sentence. If the question remains answerable, it is considered weak again and discarded. If it is no longer answerable, it is upgraded to a strong candidate.

\subsection{Cross-Model Validation}

All strong questions, including those that were already strong after the first test and those strengthened successfully after distractor regeneration, are then re-evaluated by a second reliable LLM. In our pipeline, this cross-check is performed using GPT-5.1 in non-reasoning mode.

This step is important because a question might appear strong only due to the limitations or biases of the first model. By requiring the item to remain difficult for a different LLM, we reduce model-specific artifacts and improve the robustness of the benchmark construction process.

\subsection{Non-Ambiguity Test With Sentence Context}

Questions that pass cross-model validation undergo a final test, which differs from the previous ones. In this stage, the LLM (DeepSeek-Chat-V3.2) receives the question, the original reference sentence, and the answer choices.

Now the model \emph{must} be able to answer correctly. If it still fails, then the item is considered ambiguous, poorly formulated, or insufficiently grounded in the reference sentence. Such items are discarded. Only questions that are both difficult \emph{without} the sentence and answerable \emph{with} the sentence are added to the benchmark.

This final stage ensures that retained questions are not only strong, but also valid and unambiguous. 

\subsection{Benchmark Composition Across Pipeline Stages}

\begin{figure}
    \centering
    \includegraphics[width=\linewidth]{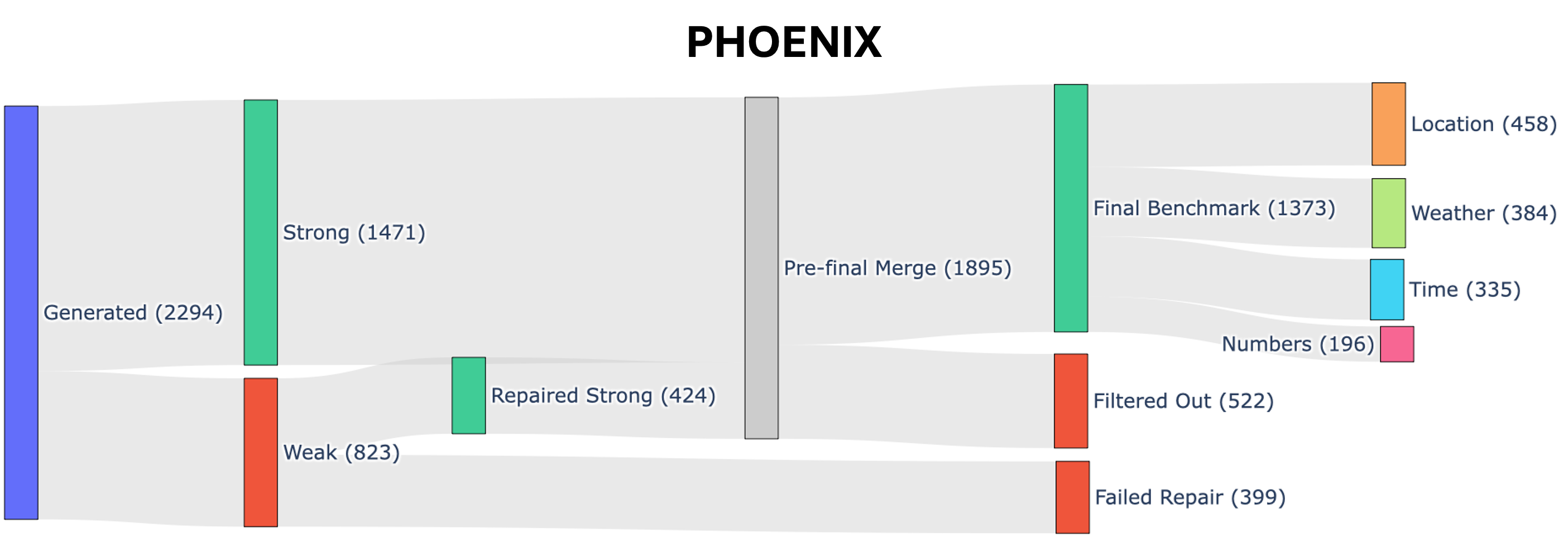}
    \includegraphics[width=\linewidth]{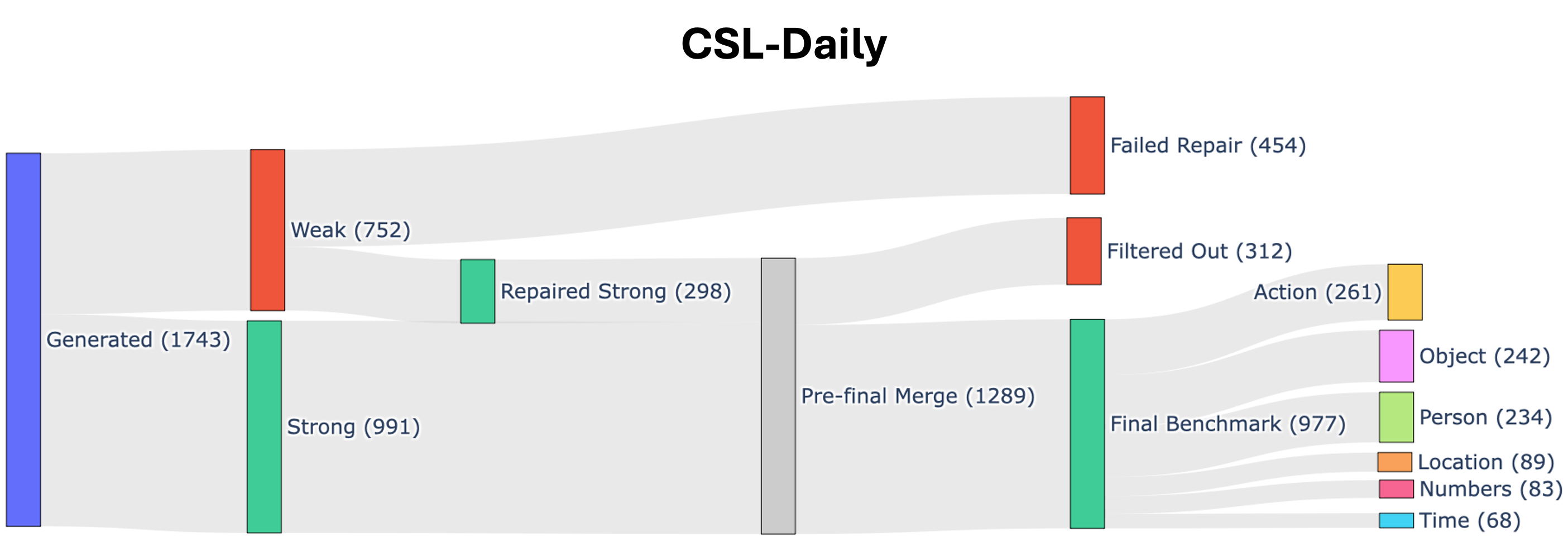}
    \caption{Benchmark composition across the main stages of data construction. \textit{Strong} indicates questions that pass the first weakness test; \textit{Repaired Strong} indicates questions that become strong after distractor regeneration (second weakness test); \textit{Final benchmark} corresponds to questions after cross-model validation and non-ambiguity filtering.}
    \label{fig:sankey}
\end{figure}

Since the proposed benchmark is automatically constructed, it is important to quantify how it evolves across the successive filtering stages. \Cref{fig:sankey} summarizes the number of questions retained at each step for both datasets and all semantic categories.

The numbers show that all the filtering stages are necessary. On PHOENIX, the initial generation stage yields 2294 questions, which drops to 1471 after the first weakness test. Among the initially weak questions that are sent to distractor repair, 424 additional items become strong after the second weakness test. The final benchmark contains 1373 questions. A similar pattern holds on CSL-Daily: generation starts from 1743 questions, 991 survive the first weakness test, 298 repaired items become strong after the second test, and the final benchmark contains 977 questions.

These results indicate that naive question generation alone would produce many weak or ambiguous items. The successive filtering stages therefore play a crucial role in transforming raw generated questions into a benchmark that is more reliable as a measure of semantic preservation.

\subsection{Human Validation}

To ensure the quality of the benchmark, we conducted a human evaluation on over 20\% of the data, involving 20 independent annotators. This setup reduces individual annotator bias and, by covering a substantial portion of the benchmark, ensures a statistically meaningful sample. Each question was assessed along multiple dimensions: \textit{clarity} (is the question well-formed and understandable), \textit{coherence} (is it consistent with the source sentence), \textit{category\_fit} (does it isolate the intended semantic aspect for the target category), \textit{correct\_answer} (is the correct answer unambiguous), \textit{distractors} (are distractors plausible yet clearly incorrect), and \textit{overall} quality. Ratings were provided on a 1--5 Likert scale. After evaluation, each item was labeled as \textit{keep}, \textit{revision}, or \textit{reject}.

This process allowed us to validate both the quality and reliability of the benchmark. The resulting average question quality was 4.69 on CSL-Daily and 4.80 on PHOENIX.

\begin{figure*}[t]
    \centering    \includegraphics[width=0.95\textwidth]{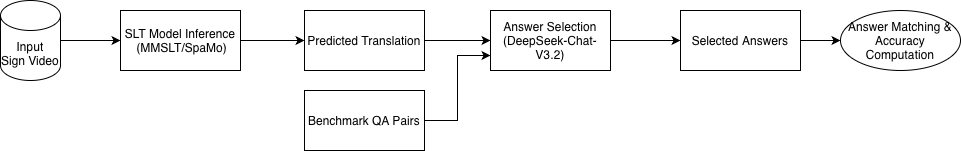}
    \caption{Overview of the model evaluation pipeline. Given an input sign video, an SLT model produces a predicted translation, which is combined with the corresponding benchmark question–answer pairs and fed to an LLM. The LLM selects an answer for each question based solely on the translation. The predicted answers are then compared with ground-truth answers to compute accuracy, measuring the model’s ability to preserve semantic information beyond surface-level similarity.}
    \label{fig:pipeline2}
\end{figure*}

\subsection{Evaluation Protocol}

Given a video, an SLT model first produces a natural language translation. This translation is then used to answer the benchmark question associated with the video. In practice, we prompt an LLM (DeepSeek-Chat-V3.2) with the generated translation and the multiple-choice question, and select the predicted answer (see \Cref{fig:pipeline2}).
We then determine whether the predicted answer matches the reference answer. Results are aggregated into overall semantic accuracy as well as per-category semantic accuracy.

This protocol complements standard translation metrics. While BLEU may remain useful as a baseline, our benchmark explicitly measures semantic preservation. The two should therefore be interpreted as complementary dimensions rather than interchangeable quantities.

\begin{table*}[t]
    \centering
    \small
    \setlength{\tabcolsep}{5pt}
    \scalebox{1}{ 
    \begin{tabular}{ll ccccc c ccccccc}
        \toprule
        & & \multicolumn{5}{c}{\textbf{Translation Metrics $\uparrow$}} & \multicolumn{8}{c}{\textbf{Sign Language Understanding Accuracy (\%) $\uparrow$}} \\
        \cmidrule(lr){3-7}  \cmidrule(lr){8-15}
        
        \textbf{Dataset} & \textbf{Model} & \textbf{B1} & \textbf{B2} & \textbf{B3} & \textbf{B4} & \textbf{R} & \textbf{Overall} & 
        \textbf{Loc.} & \textbf{Num.} & \textbf{Time} & \textbf{Weath.} & \textbf{Act.} & \textbf{Obj.} & \textbf{Pers.} \\
        
        \midrule
        
        \multirow{2}{*}{\textbf{PHOENIX}} 
        & Qwen & 1.95 & 0.20 & 0.04 & 0.01 & 3.46 & 5.52 & 4.59 & 6.12 & 4.47 & 5.99 & -- & -- & -- \\
        
        & Gemini & 4.34 & 2.03 & 1.01 & 0.61 & 8.70 & 7.36 & 6.11 & 6.12 & 9.25 & 7.81 & -- & -- & -- \\
        & MMSTL & 48.92 & \textbf{38.12} & \textbf{30.79} & \textbf{25.73} & \textbf{47.97} & \textbf{75.24} & \textbf{67.69} & \textbf{77.55} & \textbf{84.18} & \textbf{75.26} & -- & -- & -- \\
        & SpaMo & \textbf{49.80} & 37.32 & 29.50 & 24.32 & 46.57 & 68.54 & 59.39 & 66.84 & 79.40 & 70.83 & -- & -- & -- \\
        
        \midrule
        
        \multirow{2}{*}{\textbf{CSL-Daily}} 
        & Qwen & 0.00 & 0.00 & 0.00 & 0.00 & 0.00 & 5.50 & 3.37 & 6.02 & 3.84 & -- & 6.13 & 4.96 & 5.15 \\
        & Gemini & 3.16 & 1.72 & 1.15 & 0.82 & 0.00 & 5.93 & 5.61 & 6.02 & 5.88 & -- & 6.51 & 4.96 & 5.15 \\
        & MMSTL & \textbf{49.87} & 36.37 & \textbf{27.29} & \textbf{21.11} & \textbf{48.92} & \textbf{56.66} & \textbf{62.92} & \textbf{25.30} & \textbf{60.29} & -- & \textbf{61.3} & \textbf{52.07} & \textbf{63.95} \\
        & SpaMo & 48.90 & \textbf{36.90} & 26.78 & 20.55 & 47.46 & -- & -- & -- & -- & -- & -- & -- & -- \\
        
        \bottomrule
    \end{tabular}
    }
    \caption{Left: Conventional translation metrics. Here, B1--B4 denote BLEU-1 to BLEU-4, and R denotes ROUGE. Right: Overall Sign Language Understanding results  (semantic accuracy) across categories: Location (Loc.), Numbers (Num.), Time, Weather (Weath.), Action (Act.), Object (Obj.), and Person (Pers.).} 
    \label{tab:merged_results}
\end{table*}

\section{Experimental setup and results} 

As shown in \Cref{tab:merged_results}, current multimodal large language models (Qwen3-VL-4B-Instruct and gemini-3-flash) without task-specific fine-tuning are not able to perform sign language translation meaningfully, achieving near-zero scores. Corresponding semantic accuracy remains at chance level, as expected from random guessing under our evaluation protocol.
However, given the rapid progress of general-purpose MLLMs, it is plausible that future models may acquire this capability even in a zero-shot setting. In this context, our benchmark provides a simple and scalable way to track their progress over time.

For the present analysis, we therefore focus on systems that achieve meaningful performance, in order to draw reliable conclusions.
Following this, we evaluate two SLT systems on PHOENIX-2014T and CSL-Daily:
\begin{itemize}
    \item \textbf{MMSTL}, evaluated on both PHOENIX and CSL-Daily.
    \item \textbf{SpaMo}, evaluated on PHOENIX.
\end{itemize}

Performance is measured using semantic answer accuracy, defined as the ratio of correctly answered questions to the total number of questions. Results are reported both overall and per category.

\subsection{Conventional Translation Metrics}

Before analyzing semantic accuracy, we report conventional translation metrics from the original model papers in order to contextualize our results. As shown in the left part of \cref{tab:merged_results}, MMSTL and SpaMo achieve broadly similar BLEU and ROUGE scores on both PHOENIX and CSL-Daily. In particular, on PHOENIX, SpaMo slightly outperforms MMSTL on BLEU-1, while MMSTL is slightly stronger on BLEU-2, BLEU-3, BLEU-4, and ROUGE. On CSL-Daily, the two systems are again close, with MMSTL slightly ahead overall.
These results show that the models are relatively close under standard overlap-based metrics, yet they exhibit more substantial differences under our semantic benchmark. This supports the claim that conventional SLT evaluation can mask meaningful differences in semantic preservation.

\subsection{Main Quantitative Results}

The right part of \Cref{tab:merged_results} summarizes the semantic benchmark results. When compared with the conventional metrics, an important pattern emerges: models that are relatively close under BLEU and ROUGE can differ much more clearly in semantic accuracy. This is precisely the kind of gap our benchmark is designed to reveal.

Three observations are immediate. First, semantic accuracy on PHOENIX is substantially higher than on CSL-Daily. MMSTL reaches 75.24\% on PHOENIX but only 56.66\% on CSL-Daily. This suggests that PHOENIX, while still useful, is semantically easier for sign language understanding, likely because its domain is narrower and more syntactically regular.
Second, on PHOENIX, MMSTL outperforms SpaMo by 6.70 absolute points (75.24\% vs. 68.54\%). This is a meaningful gap given that the two systems are much closer under conventional BLEU and ROUGE scores. The result indicates that semantic preservation is not uniformly captured by standard translation metrics and that different architectures may trade off lexical fluency and factual precision differently.
Third, the overall numbers confirm that even state-of-the-art SLT models leave substantial room for improvement. A semantic accuracy in the 68--75\% range on PHOENIX implies that roughly one out of every four to three benchmark questions is still answered incorrectly by the translated output.

\subsection{Category-Level Analysis on PHOENIX}

\Cref{tab:merged_results} provides also a fine-grained comparison on PHOENIX. The most striking result concerns numbers. MMSTL achieves 77.55\% while SpaMo reaches only 66.84\%, a gap of 10.71 points. This indicates that preserving exact numerical information remains a major challenge in SLT systems. Since numbers often encode temperatures, quantities, or dates, such mistakes are semantically costly even when the remainder of the sentence is fluent.
For location, the gap is also large: 67.69\% for MMSTL versus 59.39\% for SpaMo. This suggests that named entities and spatial references are another fragile point in current systems. In domains such as weather and news, errors in place names can materially alter the meaning of the translation.
The time category is the strongest for both systems, especially MMSTL, which reaches 84.18\%. This likely reflects the recurring temporal structure of PHOENIX forecasts. Even so, SpaMo remains nearly five points behind MMSTL.
For weather condition, MMSTL still leads by 4.43 points.
Overall, the category-level analysis shows that semantic difficulty is unevenly distributed. The largest failures occur in categories that require exact factual preservation rather than broad topical fluency.

\subsection{Category-Level Analysis on CSL-Daily}

Results on CSL-Daily, \cref{tab:merged_results}, are consistently lower than on PHOENIX. The strongest category is person at 63.95\%, followed closely by location at 62.92\% and action at 61.30\%. Time reaches 60.29\%, while object drops to 52.07\%. The most severe weakness is again numbers, which falls to only 25.30\%.

This result suggests that numerical content is not only difficult in PHOENIX, but becomes dramatically more challenging in a broader and less syntactically regular dataset. More generally, the CSL-Daily results reinforce the claim that semantic evaluation is strongly dataset dependent. Models that appear robust in a constrained domain may degrade substantially when tested on more diverse semantics.

\section{Discussion}
\label{sec:discussion}

Our analysis highlights several key insights about semantic evaluation in SLT.
First, surface-level translation quality should not be conflated with semantic understanding. The benchmark reveals errors that overlap-based metrics such as BLEU can easily miss. Indeed, systems with similar BLEU or ROUGE scores may exhibit substantially different levels of semantic accuracy, showing that these metrics alone are insufficient to assess whether a model preserves the underlying meaning. Second, exact factual content remains a bottleneck. Categories such as numbers and locations are particularly effective at exposing semantic failures, suggesting that current SLT models struggle to faithfully preserve precise information even when producing fluent translations.
Third, dataset complexity plays a significant role. Higher scores on PHOENIX should not be interpreted as evidence that semantic understanding is solved, but rather as a reflection of the dataset’s relative regularity compared to more diverse benchmarks such as CSL-Daily.

These findings have implications for both evaluation and model development. From an evaluation perspective, conventional metrics should be complemented with methods that explicitly measure semantic preservation. While BLEU and related scores remain useful as coarse indicators of textual similarity, they cannot capture whether a model preserves the facts that matter, and may mask substantial differences in semantic accuracy across systems.
From a modeling perspective, the results suggest that SLT systems require stronger mechanisms to preserve critical factual details.
From a benchmark construction perspective, our analysis highlights the importance of the proposed pipeline. A substantial portion of initially generated questions is filtered out during validation, indicating that naive generation alone would introduce many weak or ambiguous items. The successive validation stages are therefore essential to obtain a reliable semantic evaluation resource.

Finally, the benchmark enables more fine-grained diagnosis of model behavior. Rather than relying on a single aggregate score, future work can investigate which semantic categories improve under different architectures, training strategies, or pretraining regimes, leading to more interpretable and informative evaluations.

While the proposed approach provides a useful perspective on semantic evaluation, it also presents some limitations. The benchmark relies on automatically generated question--answer pairs, which, despite the validation pipeline, may still introduce occasional noise or ambiguity. In addition, our experiments cover a limited set of models and datasets; extending the evaluation to a broader range of SLT systems would further strengthen the generality of the findings. Finally, the proposed evaluation focuses on targeted factual preservation and is therefore intended to complement, rather than replace, existing metrics capturing other aspects of translation quality.

\section{Conclusion}
\label{sec:conclusion}

We presented a question-based benchmark framework for semantic evaluation of Sign Language Translation. The framework measures whether a model prediction preserves the answers to targeted semantic questions derived from reference sentences, thereby shifting evaluation from surface overlap to semantic faithfulness.

Experiments on PHOENIX-2014T and CSL-Daily show that current SLT systems still exhibit substantial semantic weaknesses, especially on exact factual content such as numbers and locations. On PHOENIX, MMSTL outperforms SpaMo by 6.7 absolute points in overall semantic accuracy, with the largest difference appearing in the numbers category. On CSL-Daily, MMSTL reaches only 56.7\%, revealing a marked increase in difficulty compared with PHOENIX.

The benchmark composition analysis further shows that the multi-stage construction pipeline is useful in practice: many initially generated questions are filtered out before the final benchmark, suggesting that careful validation is necessary to obtain reliable semantic evaluation items.

Overall, our results suggest that strong translation metrics do not necessarily imply genuine understanding. We argue that future SLT research should evaluate models not only by how fluent their outputs look, but also by how faithfully they preserve meaning.

\section{Acknowledgements}   
This work was done during an Erasmus+ Traineeship at University of Granada. We also thank the Google Research Scholar Programme 2021. Díaz acknowledges TSI-100927-2023-1 Project (Transformation and Resilience Plan from the EU NextGen through the Ministry for Digital Transformation and the Civil Service), and Grants PID2023-149128NB-I00 and PID2023-150070NB-I00 funded by MICIU/AEI /10.13039/501100011033 and by ERDF, EU. Baraldi acknowledges the EU Horizon project “ELLIOT - European Large Open Multi-Modal Foundation Models For Robust Generalization On Arbitrary Data Streams” (No. 101214398), and the EuroHPC JU project “MINERVA” (GA No. 101182737).
{
    \small
    \bibliographystyle{ieeenat_fullname}
    \bibliography{main}
}


\end{document}